# Multi-Plane Neural Radiance Fields for Novel View Synthesis

Youssef Abdelkareem[1], Shady Shehata[2], Fakhri Karray[1,2]

[1]University of Waterloo, Waterloo, Canada
yafathi@uwaterloo.ca, karray@uwaterloo.ca
[2]Mohamed bin Zayed University of Artificial Intelligence, Abu Dhabi, United Arab Emirates
shady.shehata@mbzuai.ac.ae

## ABSTRACT

*Novel view synthesis is a long-standing problem that revolves around rendering frames of scenes from novel camera viewpoints. Volumetric approaches provide a solution for modeling occlusions through the explicit 3D representation of the camera frustum. Multi-plane Images (MPI) are volumetric methods that represent the scene using front-parallel planes at distinct depths but suffer from depth discretization leading to a 2.D scene representation. Another line of approach relies on implicit 3D scene representations. Neural Radiance Fields (NeRF) utilize neural networks for encapsulating the continuous 3D scene structure within the network weights achieving photorealistic synthesis results, however, methods are constrained to per-scene optimization settings which are inefficient in practice. Multi-plane Neural Radiance Fields (MINE) open the door for combining implicit and explicit scene representations. It enables continuous 3D scene representations, especially in the depth dimension, while utilizing the input image features to avoid per-scene optimization. The main drawback of the current literature work in this domain is being constrained to single-view input, limiting the synthesis ability to narrow viewpoint ranges. In this work, we thoroughly examine the performance, generalization, and efficiency of single-view multi-plane neural radiance fields. In addition, we propose a new multiplane NeRF architecture that accepts multiple views to improve the synthesis results and expand the viewing range. Features from the input source frames are effectively fused through a proposed attention-aware fusion module to highlight important information from different viewpoints. Experiments show the effectiveness of attention-based fusion and the promising outcomes of our proposed method when compared to multi-view NeRF and MPI techniques.*

## KEYWORDS

*Novel View Synthesis, Neural Radiance Fields (NeRF), Multi-plane Images, Volumetric Rendering, Computer Vision*

## 1. INTRODUCTION

The topic of view synthesis has caught the attention of many researchers in recent years due to its applications in fields like telepresence, virtual reality, etc. It involves rendering novel views for a particular scene using only discrete views as input. Early approaches rely on interpolation between light fields for novel view generation [1, 2] which fails to render occluded areas. Volumetric approaches offer a superior ability to handle occlusions by explicitly representing the 3D structure of the scene with different distributions [3, 4, 5, 6]. Multi-plane images (MPI) are one of the possible representations used for volumetric novel view synthesis [7, 8]. The drawback is having a sub-optimal 2.5D scene representation due to discrete depth sampling of the planes.

Another domain of approaches studies implicit neural representations to carry out novel view synthesis using neural networks and differentiable rendering. NeRF [9] encapsulated the radiance field of the scene using a Multi-Layer Perceptron (MLP). They showed impressive results for rendering synthetic and real-world scenes from a large range of novel views. The downside is that the network is trained once per scene, hence requiring full re-training procedures for each novel scene. In addition, it requires a large number of input views to get satisfactory results and the rendering quality degrades as the number of input views decreases. Generalizable NeRF methods [10, 11] offer a solution by conditioning the MLP on spatial features extracted from the input images to generalize to novel scenes.

MINE [12] is a recent approach that proposes merging the concepts of generalizable neural radiance fields [9, 11] and multi-plane images [7] in order to carry out both novel view synthesis and dense depth estimation using only one single input view while having a continuous 3D representation of the depth of the scene and requiring no per-scene optimization. The downside is that single-view settings hinder the ability to synthesize a wide range of novel views.

In this paper, we analyze the capabilities and boundaries of single-view multi-plane neural radiance fields for novel view synthesis. This is done through a detailed technical analysis of MINE [12] which showcases the performance of the method on challenging datasets and evaluates the impact of the NeRF modules on the quality of the results. In addition, we evaluate the generalization boundaries of the method by testing on novel scenes not seen during the training of different datasets. The efficiency of the method is also quantifiably evaluated against baseline NeRF approaches. Lastly, we propose a multi-view and multi-plane neural radiance field architecture, denoted as MV-MINE. The aim is to explore the effect of additional information seen from different views and whether they can contribute to high-quality results compared to state-of-the-art multi-view MPI [8] and NeRF methods [11, 10]. Our experiments demonstrate the potency of attention-based fusion and the promising results of our suggested architecture when compared to state-of-the-art multi-view NeRF techniques.

Our contributions are summarized as follows:

- We provide in-depth technical analysis on the performance, generalization, and efficiency of single-view multi-plane neural radiance fields for novel view synthesis.
- We propose, MV-MINE, an architecture merging between generalizable neural radiance fields and multi-plane images with a multi-view input setting.
- We propose an attention-based feature fusion module for effectively aggregating multi-view input.

## 2. RELATED WORK

This section will discuss the background work for novel view synthesis which is categorized into explicit and implicit 3D representations for view synthesis along with their possible combination.

### 2.1. Explicit 3D Representations

Initial work for novel view synthesis utilized the concept of light fields [1, 13, 14] which parameterizes the radiance as a 4D function of position and direction and carries out interpolation between input views to generate the target views. However, they are unable to effectively model the occlusions present in the scene. Volumetric approaches aim towards learning explicit representations of the camera frustum, which opens the door for modeling occluded regions and non-Lambertian effects. The representations include 3D voxel grids [3, 15], textured meshes [4, 16], point clouds [5], layered depth images (LDI) [6, 17] and multi-plane images (MPI) [18, 19, 8]. MPI approaches represent the scene as a set of discretized RGB-$\alpha$ front-parallel planes representing the elements of the scene at different depths. Most of the MPI approaches rely on multi-view input for novel view prediction. A recent approach [7] proves the potential of utilizing MPI in the single view setting for high-quality view synthesis. They estimate the planes using a deep CNN and introduce a scale-invariant synthesis approach to solve the scale ambiguity problem for single-view settings. Although the results were impressive, the planes are predicted at discrete depths which constrains the ability to continuously model the 3D space at any depth value.

## 2.2. Implicit 3D Representations

Another domain of approaches aims to represent the whole 3D space using a neural network architecture to act as an implicit 3D representation. Some methods [20] require only supervision from RGB images by utilizing differentiable rendering techniques but suffer from artifacts in rendered images of complex structures. A recent state-of-the-art method for view synthesis called NeRF [9] uses the concept of neural scene implicit representation for modeling the radiance field. Specifically, they utilize a Multi-layer Perceptron (MLP) which accepts a 5D input representing the 3D location of a point in a scene and the viewing direction and outputs the volume density and RGB color. This enables the encapsulation of the whole continuous 3D space of a scene inside the MLP weights. For each pixel in the input view images, a ray is transmitted across the 3D space with respect to the camera, and 3D points are sampled. Each 3D point along with the ray viewing direction passes through MLP to produce the radiance and density of the point.

The limitations of per-scene training and the high number of input views needed to employ the aforementioned NeRF method and its extensions [21, 22] make them ineffective in real-world applications. Using CNNs to predict pixel-aligned features from the input frames, generalizable NeRF approaches [10, 11, 23] employ these features together with the positional vectors to query the MLP. With just sparse views as input, this allowed novel view synthesis for scenes that were not seen during training.

## 2.3. Combination of Implicit & Explicit Representations

Recently, Multi-plane Neural Radiance Fields (MINE) [12] were proposed as a combination of implicit and explicit representations for novel view synthesis. They utilize an encoder-decoder architecture, shown in Figure 1, to predict front-parallel planes consisting of 4D radiance fields (RGB and volume density) [9] for each pixel. They sample the planes at arbitrary depth values throughout the training allowing the method to possess continuous representations of the depth dimension. This is followed by homography warping [7] and volumetric rendering [9] to render the target frame. MINE [12] proves the promising capability of marrying the concepts of NeRF with multiplane volumetric representation for high-quality synthesis results. However, being limited to a single-view setting, the method is constrained to a narrow viewing direction angle range. In this paper, we provide an in-depth technical analysis of the capabilities of single-view multi-plane neural radiance fields and propose possible solutions to extend it to multi-view settings for better synthesis ability.

## 3. METHODOLOGY

Our methodology mainly tackles the technical analysis of single-view multi-plane neural radiance fields (MINE) [12] for novel view synthesis. We additionally explain the proposed architecture, MV-MINE, to handle multi-view input for multi-plane radiance fields.

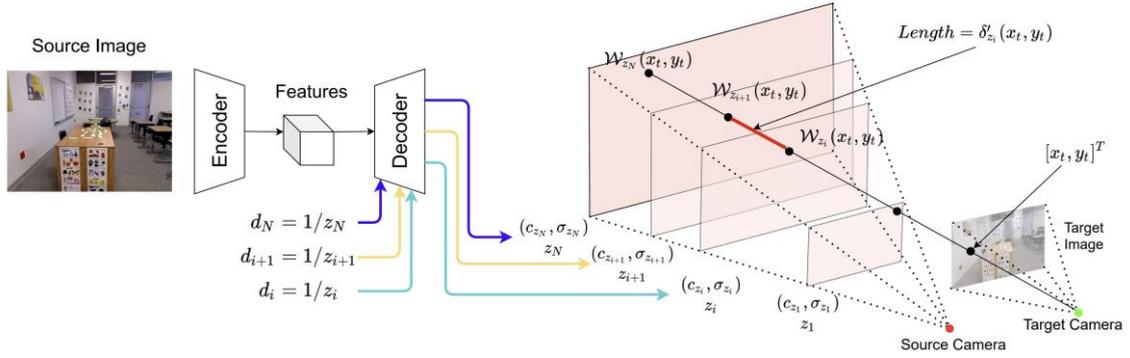

Figure 1: Full architecture of the single-view multi-plane neural radiance field [12] architecture.

### 3.1. Technical Analysis Methodology

We aim to assess three main aspects of the single-view MINE [12] architecture which are grouped into the following categories: Performance, Generalization, and Efficiency. This section will be divided based on the three categories.

#### 3.1.1. Performance

We train the network on the ShapeNet dataset [24] which is a challenging dataset used by various state-of-the-art generalizable NeRF methods [10, 11] to assess their degree of generalization through various distribution of objects in training and testing. Additionally, we carry out ablation studies to test the impact of some NeRF [9] concepts on the results of MINE. For each pixel in the target image, a 3D ray $r$ is projected into the scene. 3D points are then sampled across the ray using a specific sampling technique. Fixed-depth sampling involves sampling points at rigid depth values across all training runs which limits the representational capacity of the depth dimension. Stratified sampling involves randomly sampling points at different depth locations across the projected rays. As points are sampled in random depth locations in every training run, the method achieves a continuous depth representation by the end of all training runs. In our ablation studies, we test the performance of MINE with fixed-depth sampling and stratified sampling. To produce the final color $\hat{C}(r)$ per ray $r$, two methods exist in the literature to fuse the predicted colors of all points $i$ sampled on the ray. Alpha compositing involves carrying out an over operation [33] to aggregate the colors $c_i$ for each point $i$ based on their alpha value $\alpha_i$, such that,

$$\hat{C}(r) = \sum_{i=1}^{N} \left( c_i \alpha_i \prod_{j=i+1}^{N} (1 - \alpha_j) \right), \quad (1)$$

On the other hand, volumetric rendering involves weighing all the RGB colors $c_i$ by the density $\sigma_i$ and the depth difference $\delta_i$ of each point $i \in [1, N]$, such that,

$$\hat{C}(r) = \sum_{i=1}^{N} (T_i(1 - \exp(-\sigma_i \delta_i)) c_i), \text{ where, } T_i = \exp\left(-\sum_{j=1}^{i-1} \sigma_j \delta_j\right), \quad (2)$$

This formulation enables a more intuitive representation of occlusions in the scene. In other words, if a 3D point $i$ is occluded by points appearing before it across the ray the transmittance $T_i$ will be low and the point will contribute less to the final color $\hat{C}(r)$ of the pixel. We carry out an ablation study to compare the effects of alpha compositing and volumetric rendering on the results of MINE.

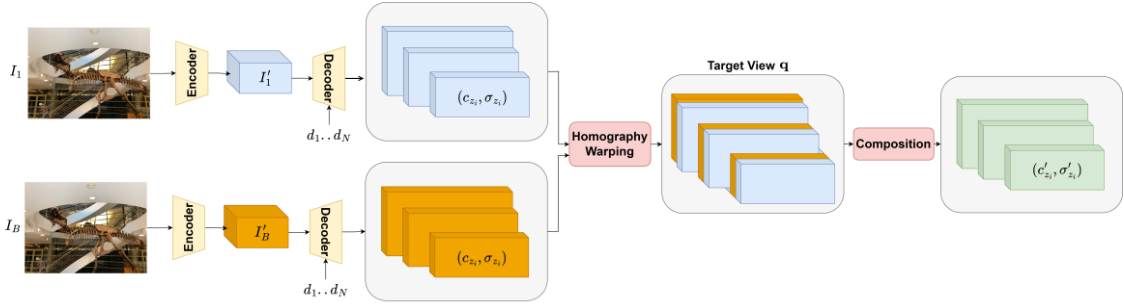

Figure 2: Full architecture of the proposed post-decoder fusion architecture design.

### 3.1.2. Generalization

We aim to validate the degree of generalization of single-view MINE to new scenes that were not seen during training. The original MINE uses an encoder-decoder network allowing the decoder to be locally conditioned on the image features extracted per pixel from the encoder. The network learns features about the scene that serves as a strong prior when presented with frames from novel scenes leading to the generalization ability. To validate this ability, we feed the model with new scenes that were not seen during training and qualitatively judge the quality of the novel views produced by the model.

### 3.1.3. Efficiency

MINE [12] is characterized to be more efficient than some of the implicit neural representation counterparts [11, 10] as it models only the frustum of the source camera, while the other synthesis methods represent the whole 3D space. During inference, MINE only produces $N$ planes corresponding to $N$ depth values from the source view to render a new view which is one single forward pass through the network. On the other hand, [11] needs to query a multi-layer perceptron for each point across a ray per pixel leading to $D \times H \times W$ forward passes through the network, where $H$ and $W$ are the height and width of the images respectively and $D$ is the number of points sampled per ray. We aim to quantify such speed-up to verify the efficiency hypothesis, while also contributing a quantitative baseline time to compare with other NeRF-variants that offer an increase in speed just like MINE.

## 3.2. Proposed Multi-view MINE Architecture

Reliance on single-view input hinders the ability of MINE [12] to render target views that are far from the source view. We explored the extension of the architecture to a multi-view input setting to leverage the rich information seen from different views for better performance on more challenging datasets, while also opening the doors to comparing with state-of-the-art multi-view synthesis methods. The following section gives an overview of the proposed architecture, MV-MINE, along with the modules used for multi-view feature fusion.

### 3.2.1. Problem Formulation

Given a synchronized set $\Omega$ of frames $I$ taken from $B$ sparse input viewpoints of a scene such that $\Omega = \{ I_1, .., I_B \}$ our target is to synthesize a novel view frame $I_q$ of the scene from a query viewing direction $\mathbf{q}$ with respect to a source view $\mathbf{s}$. Each input viewpoint $b$ is represented by the corresponding camera intrinsics $K$, and camera rotation $R$ and translation $t$, where $b = \{K_b, [R_b|t_b]\}$. For each input frame $I_w \in \mathbf{R}^{H \times W \times 3}$ with height $H$ and width $W$, we extract a multi-scale feature pyramid using a ResNet50 [25] encoder network, pre-trained on ImageNet. The operation is carried out for all input views $b$ in $\{1, .., B\}$ to produce the multi-scale feature planes for each view, defined as $I'_b \in R^{H_b \times W_b \times C_b}$.

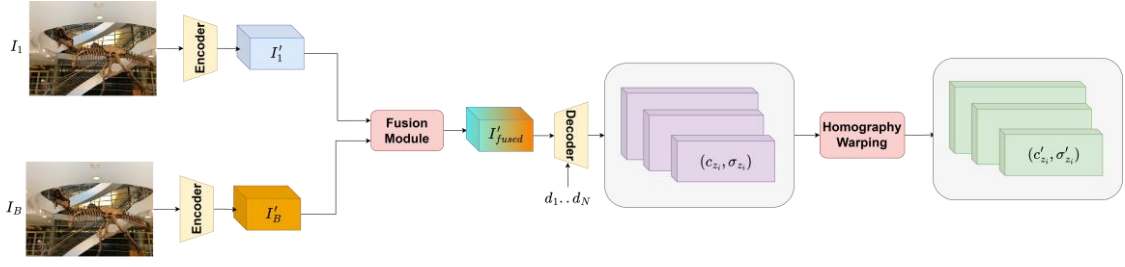

Figure 3: Full architecture of the proposed Pre-Decoder Fusion architecture design.

Similar to MINE [12], a decoder network with Monodepth2 [26] architecture takes the encoded feature maps and a disparity value $di = 1/z_i$ to produce the radiance field plane $(c_{z_i}, \sigma_{z_i})$, where $c_{z_i}, \sigma_{z_i}$ represent the color and volume density at depth $z_i$, respectively. Homography warping is then utilized to retrieve the radiance field plane $(c'_{z_i}, \sigma'_{z_i})$ at the target camera **q**. Lastly, volumetric rendering uses the predicted volume densities to aggregate the colors at different depth values producing the final target image $I_q$.

We experiment with different architecture designs to fuse the multi-view image feature planes $I'_{1..B}$. The designs include doing the fusion before or after the decoder network. We discuss both designs in the following sections.

### 3.2.2. Post-Decoder Fusion

Figure 3 shows the full post-decoder fusion architecture design. For each view $b$, the multi-scale feature planes $\{I'_b\}$ are passed along with $N$ disparity values retrieved with stratified sampling [9] to produce $N$ radiance field planes $(c^b_{z_i}, \sigma^b_{z_i})$ at different depth values. We then warp the radiance field planes from each source view to the target view using homography warping producing a set of planes $(c'^{1:B}_{z_i}, \sigma'^{1:B}_{z_i})$ aligned with the target camera frustum. To compose the radiance field planes, we can carry out basic averaging across all views such that $(c'_{z_i}, \sigma'_{z_i}) = \frac{1}{B}\sum_b (c'^b_{z_i}, \sigma'^b_{z_i})$. However, such formulation could lead to hallucinations as equal weight is given to all input views. To overcome this, we experiment with doing weighted averaging based on the distance between the source view $b$ and the target view $q$ giving higher weight to views that are closer to the target view.

### 3.2.3. Pre-Decoder Fusion

Compositing the radiance field planes after passing through the decoder for each input view is considered highly inefficient. Specifically, the decoder is invoked $N \times B$ times. A more efficient solution would fuse the multi-view feature planes before passing through the decoder leading to $N$ decoder invocations instead. We propose two fusion modules to aggregate the multi-view feature planes $I'_{1:B}$ with respect to a source view **s**. The fused multi-view features $I'_{fused}$ are then passed to the decoder to predict the radiance field planes.

*Fixed View Fusion Module.* In this module, we assume that the architecture accepts a fixed number of $B$ input views. We start by concatenating each input feature plane with their corresponding viewing direction $b_{1:B}$. All feature planes are then concatenated and passed through channel-wise fusion layers $Conv_{1\times 1}$, composed of $1 \times 1$ convolution layers with non-linear activation, to fuse the multi-view features per pixel. This is followed by $3 \times 3$ convolution $Conv_{3\times 3}$ for learning spatially fused features. The final fused features are derived by adding the source view **s** features, such that,

$$I'_{fused} = Conv_{3\times 3}\big(Conv_{1\times 1}([I'_1; \gamma(b_1)] \oplus .. \oplus [I'_B; \gamma(b_B)])\big) + I'_s \quad (3)$$

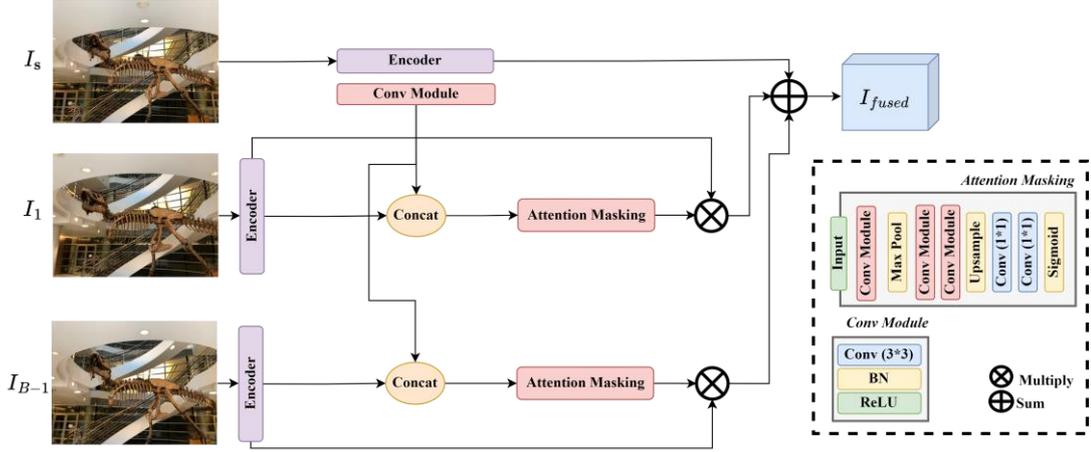

Figure 4: Full architecture of the proposed view-agnostic attention module. ($K * K$) denotes a convolution layer with $K * K$ filter size.

*Attention-based View-agnostic Fusion Module.* To increase the flexibility of our architecture with multi-view input, we propose an attention-based fusion module that accepts an arbitrary number $B$ of input views throughout training and inference. Figure 4 shows the architecture of the module. Each input view feature $I'_{1:B-1}$ is concatenated with generated source view features and passed through a soft-attention masking module. To create a soft mask, the input is down-sampled using max pooling to widen the receptive field, then the features are refined using residual units, up-sampled to their original size, and the mask is normalized to the $[0-1]$ range using a sigmoid function. The learned attention mask highlights areas of the input views that contain complementing features with respect to the source view. Input view features are multiplied by their soft mask and added to the source view features generating the final fused features $I'_{fused}$.

## 4. EXPERIMENTAL RESULTS

### 4.1. Metrics

We follow LLFF [8] in their choice of quantitative metrics in all experiments which are: LPIPS (lower the better), SSIM (higher the better), and PSNR (higher the better).

### 4.2. Technical Analysis Experiments

We present the experimental details and results of the technical analysis discussed in Section 3.1 in terms of performance, generalization, and efficiency.

#### 4.2.1. Performance

Regarding performance, we present the experimental setup and results used for training on the ShapeNet dataset [24], and the ablation studies made.

**Setup**

*Training on ShapeNet.* We train MINE on specific subsets of the ShapeNet dataset [24] to have a fair performance comparison with pixelNeRF [11] which is a generalizable single view NeRF method. Specifically, we focus on using the Category Agnostic ShapeNet experiments [11] which train on single-view images of 13 categories of objects. Each category has multiple objects and each object has 24 views. Following [11] we sample one random view for training and 23 other views as target views. The train-test split is composed of 156,877 and 45,586 source and target pairs for training and validation respectively. We trained on 4 V100 GPUs with batch size 4 and a 0.001 learning rate for the encoder and decoder. Training for one epoch takes about 6 hours and validation takes about 3 hours.

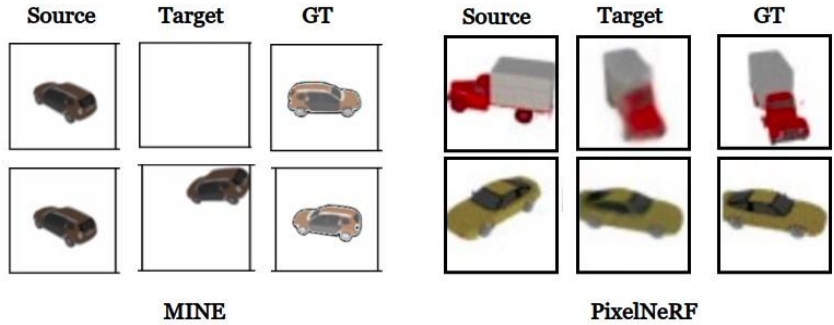

Figure 5: Output of MINE after training on ShapeNet [24] using the same preprocessing used by pixelNeRF [11]. "GT" denotes the ground truth target view," Target" denotes the output target view, and "Source" denotes the input view to the network. Distortion in GT of MINE is due to normalizing the images by 0.5.

*Effect of Continuous Depth & Volumetric Rendering.* The continuous depth reconstruction proposed by NeRF [9] allowed MINE [12] to generalize the discretized depth representation of MPI [7]. We verify this hypothesis by training on the LLFF [8] dataset from scratch with the fixed depth sampling approach from MPI [7] and the stratified sampling approach from NeRF [11]. In addition, using the volumetric rendering technique applied by NeRF [11] instead of alpha compositing [7] is one of the factors contributing to enhancing the results of MINE [12]. To verify that, we train on the LLFF dataset with both volumetric rendering and alpha compositing. The LLFF dataset [8] contains real-world images taken by phone camera at views lying in an equally spaced grid of a specific size. There are 8 scenes available with each scene having around 20-50 views available. The scenes available are of the following objects: fern, flower, fortress, horns, leaves, orchids, room, and trex. During the training, for each view in the scene, a random view is taken as the target view. The sparse disparity loss is included, and the scale is calculated using 3D point clouds estimated for the images using COLMAP [27, 28]. Training was done on 4 V100 GPUs and took around 4 hours. We used a decaying learning rate starting at 0.001 and decaying by 0.1 every 50 epochs for 200 epochs and a batch size of 2.

**Results**

*Training on ShapeNet.* We carried out a qualitative analysis to check the plausibility of results returned by MINE compared to pixelNeRF [11] with single-view input on ShapeNet [24], shown in Figure 5. The first row shows that MINE failed to render the target object within the boundaries of the image plane since the target viewing direction is very far from the source viewing direction. In the second row, the object was rendered within the image plane and the structure of the car was retained appropriately since the two viewing directions are closer, in this case, however, the car location is still inaccurate. On the other hand, pixelNeRF is able to correctly render the target view object in an accurate location within the image plane regardless of how far the source and target views are.

*Effect of Continuous Depth & Volumetric Rendering.* Table 1 shows the results after training MINE on LLFF with fixed disparity taken at equally spaced locations, with a random stratified sampled disparity in each training step, and with volumetric rendering and alpha compositing for aggregating the colors from the radiance field planes. It can be seen that the usage of stratified sampling did not enhance the results, yet the fixed disparity yielded slightly better performance on all metrics. However, the usage of volumetric rendering led to significantly better results than alpha compositing.

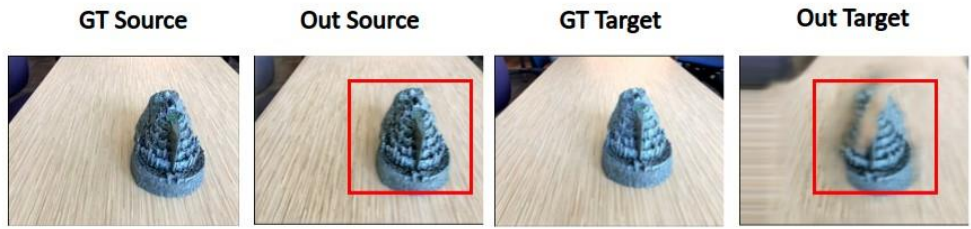

Figure 6: Output of MINE after training it on 7 LLFF [8] categories and evaluating on the fortress scene. "GT" denotes ground truth and "Out" denotes output of model.

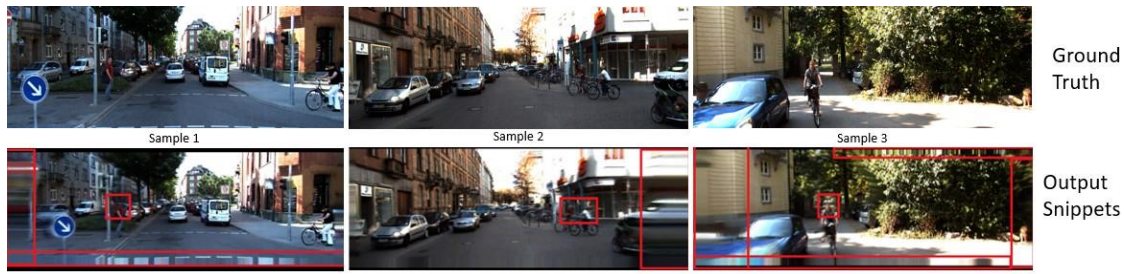

Figure 7: Global problems encountered in the KITTI Raw [30] generalization experiment.

### 4.2.2. Generalization

Regarding generalization, we present the experimental setup and results of evaluating MINE [12] on novel scenes from the LLFF [8] and KITTI Raw [29] datasets.

**Setup**

*Generalization on LLFF.* In this experiment, we leave out the" fortress" scene from the LLFF dataset during training and evaluate it. This setting is considered challenging as the novel scene differs highly from the scenes seen during training. We follow the same experimental setup of the ablation studies mentioned in Section 4.2.1.

*Generalization on KITTI Raw.* We utilize samples of scenes from the KITTI Raw [29] dataset which were not seen during training (specifically scenes dated 2011_09_26 scenes 0104, 0106, 0113, and 0117). The model is tested on each image in the scenes individually. The GPU used for this experiment is NVIDIA GTX1070 8GB.

**Results**

*Generalization on LLFF.* The results of the generalization experiment on the fortress scene on the LLFF dataset [8] are shown in Figure 6. In the second column, it is clear that the model was successful in rendering the geometric structure of the source image accurately. However, regarding the target novel views in the fourth column, it is clear the model failed to render the geometric structure of the whole object properly showing a lot of distortions.

*Generalization on KITTI Raw.* The results of testing the generalization on KITTI Raw [29] made us consider two main divisions of the problems encountered, the division of global problems which are visible in almost all of the pictures tested, and local problems which are visible in specific frames of the scenes. The first global problem is edge distortion where the edges of the videos while moving along the z-axis are highly distorted. This happens due to duplicating the edge pixels to in-paint parts which were occluded in the source image, as visible in Figure 7.

Table 1: Results of comparing rendering time per frame for pixelNeRF[11] and MINE[12].

| Method | GPU Time | CPU Time |
|---|---|---|
| Original paper (32 Planes) | 0.77 s | 8.43s |
| pixelNeRF [11] (32 Coarse Points) | 1.24 s | 15.45s |

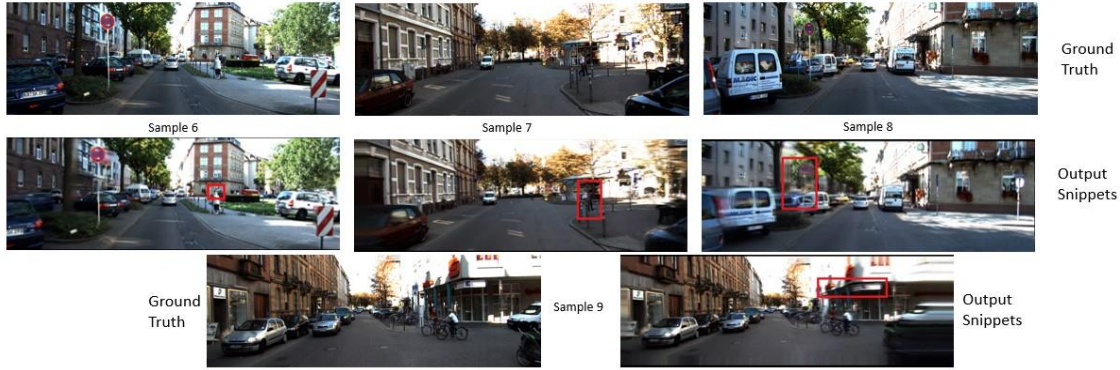

Figure 8: Local problems encountered in the KITTI Raw [30] generalization experiment.

Another global problem is rendering pixels where an object is behind another object which is visible clearly in Figure 7 samples 1-3. Specifically, in sample 1, it is visible in the sign at the front, when trying to render the car behind it. In sample 2, it is visible on the right of the motorcycles, where motorcycles are getting distorted and rendered unsuccessfully due to small barriers in front of them. In sample 3, it is visible when looking at the car on the right, when the camera moves the car shape changes. Lastly, the heads of pedestrians show large distortions as visible in Figure 7 samples 1 and 4, or have a ghost-like effect, as seen in samples 2 and 3. Locally, we highlight areas in Figure 8 where pedestrians, traffic signs, and buildings suffer from splitting distortions, ghost-like effects, and the incorrect representation of the geometric structure.

### 4.2.3. Efficiency

Regarding efficiency, we present the experimental setup and results of comparing the inference speed of MINE [12] and pixelNeRF [11].

**Setup**

We fixed the input frame shape to (128, 128) and the number of planes in MINE to 32 to be the same as the number of points sampled per ray in pixelNeRF. We used the pre-trained models and the code published for both pixelNeRF and MINE to run the experiments. The GPU used for the experiment is NVIDIA GTX1070 8GB, and the CPU is Intel(R) Core (TM) i7-6700K CPU @ 4.00GHz with 8 cores and 32 GB RAM. To obtain an accurate time per frame, we ran 150 frames and got the average time per frame.

**Results**

Table 1 presents the results of this experiment. We were able to validate that MINE is more efficient in inference than pixelNeRF [11]. Particularly, MINE renders a single (128,128) target frame in 0.77 seconds on GPU, while pixelNeRF takes 1.24 seconds which is approximately 38% speed enhancement. Regarding CPU, MINE shows 45% enhancement over pixelNeRF.

Table 2: Quantitative comparison of the performance of the proposed multi-view fusion modules using 5 input views and MINE using a single input view.

| Method | LPIPS ↓ | SSIM ↑ | PSNR ↑ |
|---|---|---|---|
| MINE [12] | 0.397 | 0.5244 | 18.12 |
| Post-Decoder Fusion (Averaging) | 0.354 | 0.601 | 19.56 |
| Post-Decoder Fusion (Weighted Averaging) | 0.298 | 0.652 | 20.43 |
| Pre-Decoder Fusion (Averaging) | 0.321 | 0.621 | 20.10 |
| Fixed-View Pre-Decoder Fusion | 0.232 | 0.761 | 24.08 |
| Attention-based Pre-Decoder Fusion | **0.223** | **0.803** | **24.43** |

Table 3: Comparison of our attention-based view-agnostic fusion module, with baseline view synthesis methods. "P" denotes per-scene optimization methods, while "G" denotes generalizable methods.

| Method | LPIPS ↓ | SSIM ↑ | PSNR ↑ |
|---|---|---|---|
| SRN (P) | 0.378 | 0.668 | 22.84 |
| NeRF (P) | 0.250 | 0.811 | 26.50 |
| LLFF (G) | 0.212 | 0.798 | 24.13 |
| pixelNeRF (G) | 0.224 | 0.802 | 24.61 |
| Ours (G) | 0.218 | 0.808 | 24.56 |

### 4.2.4. Discussion

Regarding the results of training on ShapeNet [24] in Section 4.2.1, we concluded that MINE is limited only to render novel views that are close to input source views, and in the current setting would fail to give 360° views of a scene like other NeRF variants [10, 11]. We believe that the reason behind that is having only a single image as input, so the model doesn't get exposed to several views to enhance its novel view prediction on far target poses. Moreover, homography warping could be another reason why the model has limited capability to render a wide range of views since the decoder is only producing a feature plane representation that is conditioned on the source image, and transforming the output planes by a large amount is considered ill-posed and would cause the distortion and incorrect results shown previously. For the LLFF generalization experiments, it could be concluded that MINE cannot generalize to areas around the edges of the image since it will need to in-paint the content of areas of the image that it hasn't seen before from the single-view input. In the output, we saw that the model does nearest neighbor interpolation in those areas instead of correctly predicting their structure and color. The method also failed to appropriately render the fortress scene due to its disparate distribution compared to the training scenes which highlights the weak generalization ability of the method.

### 4.3. Multi-View MINE Experiments

Our experiments in this section focus on evaluating the performance of the proposed architecture designs for MV-MINE, described in Section 3.2, and comparing them against baseline NeRF methods. We discuss the experimental setup, while also presenting the results both quantitatively and qualitatively.

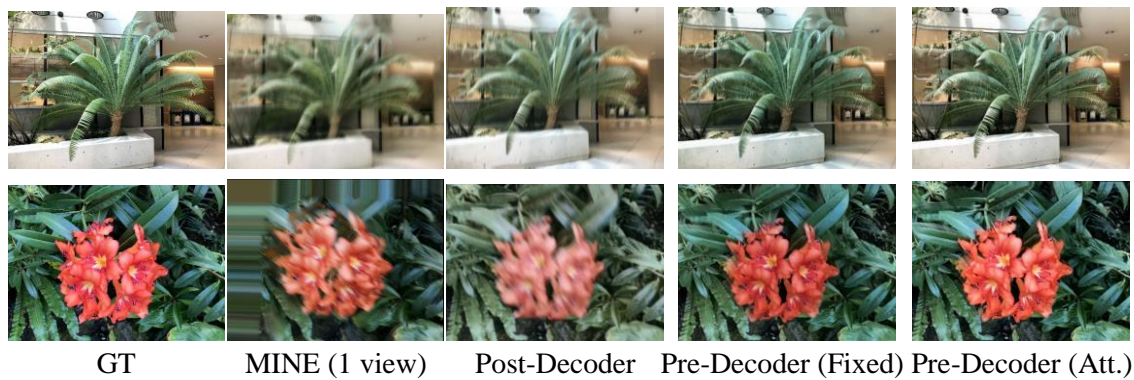

GT      MINE (1 view)     Post-Decoder    Pre-Decoder (Fixed)   Pre-Decoder (Att.)

Figure 9: Comparison of the proposed multi-view fusion modules. We include the original MINE [12] method operating with single input views. All fusion modules were tested with 5 input views.

### 4.3.1. Experimental Setup

The experimental setup involves the training and testing details and the datasets used in all experiments. The experiments are split into an evaluation of the proposed modules and a comparison with baseline methods.

*Comparison of fusion techniques*. We train the fusion modules on all the scenes of the LLFF [8] dataset. Validation is done on unseen target views. The fixed-view pre-decoder module was trained and evaluated on 5 input views, while other modules used a range of 3-7 input views for training and were evaluated on 5 input views for a fair comparison.

*Comparison with baseline methods*. Regarding per-scene methods, we evaluate our approach against NeRF [9] and SRN [31]. Regarding generalizable methods, we include pixelNeRF [11] in our baselines. In addition, we provide the results of LLFF [8] as an MPI method. Our proposed method and pixelNeRF were both trained on a collection of the LLFF [8], Spaces [32], IBRcollected [23], and RealEstate-4k [19] datasets. Training samples for each epoch are drawn with the following probabilities 0.4, 0.15, 0.35, and 0.1 respectively. Evaluation is done on novel target views of the LLFF dataset. NeRF was trained on each scene of the LLFF dataset separately.

### 4.3.2. Results

*Comparison of fusion techniques.* Table 2 presents the performance of the original MINE method with single view inputs along with our proposed fusion modules operating on 5 input views. It could be seen that the post-decoder fusion with averaging leveraged multi-view information to enhance results compared to single-view MINE. Introducing weighted averaging led to better utilization of features from close views and significantly enhanced results on all metrics. Implicit feature aggregation introduced in the fixed-view pre-decoder fusion notably elevated the performance. Lastly, shifting to view-agnostic attention-aware fusion shows the best overall performance on all metrics. This validates the impact of the learned soft masks in highlighting important features in the input views with respect to the source view. Qualitatively, it could be seen in Figure 9 that MINE suffers from strong hallucinations around image borders. The post-decoder module solves that issue yet still contains strong blur artifacts. The pre-decoder modules show the best synthesis quality, especially with the attention-aware module in terms of lighting and colors.

*Comparison with baseline methods.* Table 3 shows the results of our attention-aware fusion module compared to the baseline view synthesis methods. Regarding per-scene methods, it could be seen that our method significantly surpasses SRN on all metrics, while performing better than NeRF on the LPIPS metric without per-scene training. Regarding the generalizable methods, we show comparable performance to both LLFF and pixelNeRF, while performing better than pixelNeRF on the LPIPS and SSIM metrics. We also introduce slight improvements over LLFF on the SSIM and PSNR metrics.

## 5. CONCLUSION

In this paper, we explored the boundaries and capabilities of the combination between neural radiance fields and multi-plane images. Specifically, we analyzed the performance, generalization, and efficiency of single-view multi-plane radiance fields (MINE) [12] through training on challenging datasets [24], doing an ablation study on NeRF concepts, evaluating unseen scenes, and providing a qualitative efficiency comparison with baseline methods [11]. Our analysis led us to the conclusion that single-view MINE can only synthesize novel views that are relatively close to the input view, while not generalizing well to novel scenes not seen during training. We also proved the superior efficiency of MINE compared to pixelNeRF [11] on GPU and CPU. Furthermore, we proposed a multi-view multi-plane neural radiance field architecture, MV-MINE, which effectively utilizes information from different viewpoints to enhance the view synthesis performance. The architecture does multi-view feature fusion using a newly proposed attention module that works on any arbitrary number of views. Our experiments showcase the effectiveness of the attention-based fusion and the promising performance of our proposed approach compared to state-of-the-art multi-view NeRF methods. We believe this paper can open the door for future work to tackle the highlighted limitations of multi-plane radiance fields and capitalize on the promising potential of the domain in both single and multi-view settings.